\pdfoutput=1

\documentclass[11pt]{article}

\usepackage{acl}

\usepackage{times}
\usepackage{latexsym}
\usepackage{amsmath}
\usepackage{amssymb}
\usepackage{tikz}
\usepackage{multirow}
\usepackage{multicol}
\usepackage[T1]{fontenc}

\usepackage[utf8]{inputenc}

\usepackage{microtype}

\title{Generalised Spherical Text Embedding}

\author{Souvik Banerjee$^1$, Bamdev Mishra$^2$, Pratik Jawanpuria$^2$, \and Manish Shrivastava$^1$ \\
  $^1$ International Institute of Information Technology, Hyderabad \\  
  $^2$ Microsoft India \\
    {\small{\tt souvik.banerjee@research.iiit.ac.in}} \\
    {\small{\tt bamdevm@microsoft.com }} \\
    {\small{\tt pratik.jawanpuria@microsoft.com }} \\
    {\small{\tt m.shrivastava@iiit.ac.in }}
}

\newcommand{\R}{\mathbb{R}}
\newcommand{\wtov}{\texttt{Word2vec}}
\newcommand{\dtov}{\texttt{Doc2vec }}
\newcommand{\jose}{\texttt{JoSE}}
\newcommand{\glove}{\texttt{GLoVE }}

\begin{document}
\maketitle
\begin{abstract}
This paper aims to provide an unsupervised modelling approach that allows for a more flexible representation of text embeddings. It jointly encodes the words and the paragraphs as individual matrices of arbitrary column dimension with unit Frobenius norm. The representation is also linguistically motivated with the introduction of a novel similarity metric. The proposed modelling and the novel similarity metric exploits the matrix structure of embeddings. We then go on to show that the same matrices can be reshaped into vectors of unit norm and transform our problem into an optimization problem over the spherical manifold. We exploit manifold optimization to efficiently train the matrix embeddings. We also quantitatively verify the quality of our text embeddings by showing that they demonstrate improved results in document classification, document clustering, and semantic textual similarity benchmark tests.\footnote{\url{https://github.com/SouvikBan/matrix_rep}}.
\end{abstract}

\section{Introduction}
Most unsupervised text embedding models are trained by encoding the words or paragraphs acquired from the training data as a feature length vector, with the assumption that they reside in the Euclidean space. Such models are ubiquitous for good reason. Aside from their efficiency, they have also proven to be very effective providing us with state of the art results in various intrinsic and extrinsic embedding evaluation tasks. \wtov \cite{mikolov2013distributed, mikolov2013efficient}, and \glove \cite{pennington2014glove} are two notable examples where word embeddings are learned in the Euclidean space and are trained to be oriented such that word vectors that appear in the same context have higher cosine similarity. Some of the most common methods of intrinsic evaluation of word embeddings include word similarity, word analogy, and compositionality. \dtov \cite{le2014distributed}, an unsupervised document embedding model generalises the training method introduced in \wtov \, to documents and achieves improved results in various downstream tasks like sentiment analysis, information retrieval and multi-class classification. There are other document embedding models like skip-thought \cite{skiptho} and infersent \cite{infersent4,sent2vec}.

The joint spherical embedding model, \jose \, as proposed in \cite{meng2019spherical}, shows that directional similarity is often more effective in tasks such as word similarity and document clustering. They show that when embeddings are trained in the Euclidean space, there is a performance gap between the training stage and usage stage of text embeddings. To bridge that gap, they propose a model which trains both words and paragraphs on a spherical space with tools from Riemannian optimization methods. The resulting embeddings are also shown to give considerably better results in word similarity, document clustering, and document classification tasks when compared with other standard models. Such application of manifold geometry has also been explored in substantial depth in works like \cite{batmanghelich-etal-2016-nonparametric, stmreis, pmlr-v32-gopal14}. There are also other notable Riemannian optimization based embedding training models like \cite{poincareglove,poincarerep} which train embeddings on the hyperbolic manifold space and uses its tree like property for better hierarchical representation of data. Hyperbolic word embeddings are also intrinsically linked with Gaussian word embeddings \cite{word2gauss} which gives a lot more insight into the geometry of word embeddings. 

However, most of these text embedding models like \jose,\, \texttt{Word2vec}, \texttt{Doc2vec}, and fastText \cite{bojanowski2017enriching,topmine} are trained with the goal of getting a single dense vector representation per word or document. These models treat both polysemous and monosemous words in the same way resulting in the most frequent meaning of the word dominating the others or the meanings getting mixed in the case of former. It is especially detrimental for documents where we use a single dense vector representation to encode information which span over several sentences, often involving multiple topics.

This paper aims to address this problem by using matrices as the mode of representation instead of vectors. Our model is the joint word and document training generative model proposed in \jose \, where we replace the cosine similarity metric with a novel metric that exploits the matrix structure of the embeddings. This robust metric takes word or document matrices of arbitrary number of columns and calculates the similarity between them. We also show that a few reshape operations allow us to reformulate the optimization problem of our model in terms of the spherical manifold optimization problem. Thus, we offer more flexibility in the way of matrix dimensions while retaining efficiency. Our choice of metric also suggests that the word, sentence, paragraph/document embeddings do not need to have the same number of columns, which has linguistic validation.

\section{Matrix Representation of Texts and Optimization Problem}

The text embeddings are represented as elements of the following set
$$\mathcal{S}(p, r) = \{ \textbf{X} \in \R^{p \times r} : ||\textbf{X}||_F=1 \},$$ where $r \leq p $ and $\|\cdot\|_F$ denotes the Frobenius norm. The Frobenius norm is the matrix norm of a $p \times r$ matrix $\textbf{X}$ defined as the square root of the sum of the absolute squares of its elements, i.e., $$||\textbf{X}||_F = \sqrt{\sum_{i=1}^{p}\sum_{j=1}^{r}x_{ij}^2}.$$

Our model design is consistent with \jose \, where it is assumed that text generation is a two-step process: a center word is first generated according to the semantics of the paragraph, and then the surrounding words are generated based on the center word’s semantics. Consider a positive tuple $(\mathcal{U}, \mathcal{V}, \mathcal{D})$ where word $\mathcal{V}$ appears in the local context window of word $\mathcal{U}$ in paragraph $\mathcal{D}$ and negative tuple $(\mathcal{V}, \mathcal{U}', \mathcal{D})$ where $\mathcal{U}'$ is a randomly sampled word from the vocabulary serving as a negative sample. We represent words $\mathcal{V}, \mathcal{U}, \mathcal{U}'$ as matrices $\textbf{V}, \textbf{U}, \textbf{N}$ which are elements of the set $\mathcal{S}(p, r_1)$ and paragraph $\mathcal{D}$ as matrix $\textbf{D}$ which is an element of the set $\mathcal{S}(p, r_2)$, where $p, r_1, r_2 > 0$. 
From a linguistic perspective, these matrices can be considered as a set of latent variables that govern the semantics of a word or a document. Each column is given some arbitrary unit of linguistic information to encode, a latent variable which contributes to the mathematical representation of a word or a document. For example, the columns of a matrix $\textbf{D}$ that represent the document $\mathcal{D}$ might encode latent variables that contain information about some topic contained in that document. Similarly, the columns of the word matrix $\textbf{U}$ might encode information about a specific context in which a polysemous word  $\mathcal{U}$ appears. We also keep the number of columns for word matrices less than or equal to the number of columns for sentence/document matrices, i.e., $r_1 \leq r_2$, so that the number of latent variables governing a word should not be more than the ones that govern a sentence or paragraph. 

\textbf{Novel metric.} To model the above mentioned linguistic representation mathematically, we define a novel similarity metric for the ambient space in which we train our matrix embeddings. The proposed metric function is a measure of similarity between two sets of latent variables (matrices) - a function analogous to the cosine similarity measure for vectors in the Euclidean space. Given two arbitrary matrices $\textbf{A}\in \mathcal{S}(p, r_1), \textbf{B} \in \mathcal{S}(p, r_2)$, we propose the similarity metric $\textbf{g} : \mathcal{S}(p, r_1) \times \mathcal{S}(p, r_2) \rightarrow \R $ as
\begin{equation} \label{eq:1}
\textbf{g}(\textbf{A},\textbf{B}) = \frac{\sum_{i=1}^{r_1}\sum_{j=1}^{r_2} a_i^\top b_j}{r_1r_2},
\end{equation} 
where  $\textbf{A} = [a_1\, a_2 \, \cdots \, a_{r_1}]$ , $\textbf{B} = [b_1\, b_2 \, \cdots \, b_{r_2}]$ , $a_i, b_j \in \R^p \, \forall i \, \in [1, 2, \cdots r_1]$ , $\forall \, j \in [1, 2, \cdots r_2]$. 

\textbf{Motivation for our similarity metric.} The metric $\textbf{g}$ (\ref{eq:1}) calculates the average of all the entries in the matrix $\textbf{A}^{\top} \textbf{B}$. The linguistic intuition behind the choice of this metric is that we want to define a metric that takes the average of dot products between all possible pairs of latent variables (columns) from each matrix. In the case of $r_1 = r_2 = 1$, $\textbf{g}(\textbf{A},\textbf{B})$ reduces to the cosine similarity metric between unit norm vectors $\textbf{A}$ and $\textbf{B}$ which is the metric used in the spherical space model of \jose. However, in the case of higher values of $r_1, r_2$, For example, let two words $\mathcal{V}_1$ and $\mathcal{V}_2$ be represented by proposed $p \times r_1$ matrix embeddings - $\textbf{V}_1=[\text{a}_1,\text{a}_2]$ and $\textbf{V}_2 = [\text{b}_1,\text{b}_2]$, where $p=1$ and $r_1=2$. The proposed similarity metric $\textbf{g}(\textbf{V}_1,\textbf{V}_2)$ is computed as 
$ {(a_1b_1+a_1b_2+a_2b_1+a_2b_2})/{4} $. Note that this is different from computing the cosine similarity which gives $(a_1b1+a_2b_2)$. Moreover, the regular cosine similarity between word and paragraph embedding matrices with unequal dimensions $(p \times r_1$ and $p \times r_2$ respectively) is not defined. On the other hand, our proposed similarity metric is still applicable.

\textbf{Modelling.} As our model has the same generative process as \texttt{JoSE}, we take the same max-margin loss function and substitute the cosine similarity metric with our new similarity metric $\textbf{g}$ where the word matrices $\textbf{U}, \textbf{V}, \textbf{N} \in \mathcal{S}(p, r_1)$ and paragraph matrix $\textbf{D} \in \mathcal{S}(p, r_2)$ with $r_1 \leq r_2$. We get the following loss, i.e., 
\begin{equation} \label{eq:2}
\begin{array}{lll}
\mathcal{L}(\textbf{V}, \textbf{U},\textbf{N}, \textbf{D}) \\
\quad \quad = \max \, (0, m - \textbf{g}(\textbf{V}, \textbf{U}) - \textbf{g}(\textbf{U}, \textbf{D}) \\
 \qquad \qquad \qquad \quad  + \ \textbf{g}(\textbf{V}, \textbf{N}) + \textbf{g}(\textbf{N}, \textbf{D})) .    
\end{array}
\end{equation} 
where $m > 0$ is the margin.

\textbf{Optimization.} 
For the purpose of optimization, matrices of different dimensions are reshaped and embedded into Riemannian spherical manifolds of different dimensions. Overall, they are combined using the Riemannian product manifold structure. Therefore, the optimization of $\mathcal{L}$ (\ref{eq:2}) is done by performing two reshape operations per iteration while training. For example, the unit Frobenius norm matrices of dimension $\R^{p \times r}$ can be reshaped into vectors of dimension $\R^{pr}$ with the unit norm. To calculate the value of our loss function (\ref{eq:2}) at every iteration and the Euclidean gradient (partial derivatives), the vectors in question are reshaped into matrices for calculating the $\textbf{g}$ values and their gradients. Subsequently, the matrices are reshaped back into vectors. We then apply the Riemannian gradient descent update rule to update the parameters \cite{meng2019spherical,AbsMahSep2008,optimcontrol,edelman54}. Note that our proposed modelling and optimization are different from just training on the spherical manifold with unit vectors and using the cosine similarity metric (which is the case in \jose). 


\section{Experiments}
To highlight the quality of our obtained matrix representations, we run the same set of evaluations as \jose \, with a relatively lower number of columns, i.e., $1 \leq r_1 \leq r_2 \leq 6$. We notice that for even higher values, the quality of our embeddings gradually decrease. Moreover, the word similarity experiment results are not added as words seemingly do not benefit from our representation directly. Indeed, the best word similarity score are obtained for $r_1=r_2=1$. Instead, we add semantic textual similarity benchmark tests to show that sentences can benefit from this matrix representation model. Unless otherwise stated, our model and \jose \, are trained for 35 iterations on the respective corpora; the local context window size is 5; the embedding dimension is kept at 100; the number of negative samples are 2. Other hyperparameters in our model are kept the same as \jose. 

\subsection{Document Clustering}
We perform document clustering on the 20 Newsgroup\footnote{\url{http://qwone.com/~jason/20Newsgroups/}} dataset using spectral clustering. Each paragraph in the dataset is separated by a new line and is considered a separate document while training. \jose \, uses K-Means and SK-Means as the clustering algorithm that assume the ambient space to be the Euclidean and the spherical space, respectively. Our non-Euclidean space with its custom metric requires a clustering algorithm that allows the freedom of using custom metric, i.e., the algorithm should be space agnostic. We found spectral clustering to suit those requirements perfectly. The four external measures used for validating the results are kept unchanged from \jose \, \cite{Banerjee2005ClusteringOT,manning2008,Steinley2004PropertiesOT}. These measures are Mutual Information (MI), Normalized Mutual Information (NMI), Adjusted Rand Index (ARI), and Purity. 
We run the clustering algorithm with our custom similarity metric as written in (\ref{eq:1}) with kernel coefficient, $\gamma= 0.001$. Table \ref{tab:clustering-table} shows quantitatively how matrix representations benefit document embeddings for clustering tasks. Keeping $r_1=1$ fixed, we see a steady increase in performance as $r_2$ is increased from 1 (the score of our baseline model - \jose \,) to 6.   
\begin{table}[t]
\caption{Evaluation results for spectral clustering of document embeddings on the 20 Newsgroup dataset for kernel coefficient, $\gamma=0.001$ ($r_1=1, r_2=1$ is \jose \, score). Document embeddings benefit from matrix representations as demonstrated by better scores for higher values of $r_2$. Here, $r_1=1$. } 
\resizebox{\columnwidth}{!}{
{\tiny
\begin{tabular}{|l|l|l|l|l|}
\hline
$r_2$, $r_1=1$ & MI & NMI & ARI & Purity \\ \hline
$r_2=1$ & 1.73 & 0.58 & 0.45 & 0.64 \\ \hline
$r_2=2$ & 1.75 & 0.59 & 0.46 & 0.63 \\ \hline
$r_2=3$ & 1.75 & 0.59 & 0.46 & 0.62 \\ \hline
$r_2=4$ & 1.77 & 0.60 & 0.46 & 0.63 \\ \hline
$r_2=5$ & 1.84 & 0.62 & 0.49 & 0.65 \\ \hline
$r_2=6$ & \textbf{1.85} & \textbf{0.62} & \textbf{0.49} & \textbf{0.67} \\ \hline
\end{tabular}
}
}
\label{tab:clustering-table}
\end{table}

\subsection{Document Classification}
Following \cite{meng2019spherical}, the document classification evaluations are ran on the following two datasets: the topic classification 20 Newsgroup dataset (which we used for document clustering as well) and a binary sentiment classification dataset consisting of 1\,000 positive and 1\,000 negative movie reviews\footnote{\url{http://www.cs.cornell.edu/people/pabo/movie-review-data/}}. The train/test split is the original split for 20 Newsgroup while for the movie review datasets, the splitting is done by randomly selecting 80\% of the data as training and 20\% as testing. The classification algorithm we use is K-NN with $k=3$ and a custom distance metric that is suitable for our space. The custom distance metric for two paragraph matrices $\textbf{U} = [u_1\, u_2 \, \cdots \, u_{r_2}]$ and $\textbf{V} = [v_1\, v_2 \, \cdots \, v_{r_2}]$ where $\textbf{U},\textbf{V} \in \mathcal{S}(p, r_2)$ and $u_i, v_i \in \R^p \, \forall i \in [1, 2, \cdots, r_2] $ is defined as 
\begin{equation} \label{eq:3}
\begin{array}{lll}
\text{dist}^2(\textbf{U}, \textbf{V}) 
= \frac{{\sum_{k=1}^{r_2}\sum_{l=1}^{r_2} (u_k - v_l)^{\top} (u_k - v_l)}}{r_2^2}.  
\end{array}
\end{equation}
The intuition for the distance metric in (\ref{eq:3}) comes from our interpretation of each individual column as encoding a latent variable governing the semantics of that specific document. A quick look at (\ref{eq:3}) tells us that the distance metric takes the square root of the average of the squared Euclidean distances between all pairs of columns formed from one matrix with another.  
Tables \ref{tab:classification-table1} and \ref{tab:classification-table2} list the Macro-F1 and Micro-F1 scores for 20 Newsgroup dataset and Movie Reviews dataset respectively for increasing values of $r_1$ and $r_2$. We again see an increase in scores for higher values of both $r_2$ and $r_1$ compared to \jose \, ($r_1$=1, $r_2$=1). 

\begin{table}[t]
\caption{F1-macro, F1-micro for 20 Newsgroup dataset classification using K-NN with K=3 ($r_1=1, r_2=1$ is \jose \, score). Increasing the value $r_2$ benefits documents embeddings in classification tasks.}
\resizebox{\columnwidth}{!}{
\begin{tabular}{|l|l|l|l|l|}
\hline
$r_1$\textbackslash{}$r_2$ & $r_2$=1 & $r_2$=2 & $r_2$=3 & $r_2$=4 \\ \hline
$r_1$=1              &  0.74, 0.74  &  0.77, 0.77   &  0.78, 0.78  &  \textbf{0.78, 0.78}  \\ \hline
$r_1$=2              &  --  &  0.76, 0.76  &  0.77, 0.77  & 0.76, 0.77   \\ \hline
$r_1$=3              &  --  &  --  &    0.76, 0.76 & 0.73, 0.74   \\ \hline
$r_1$=4              & --  & --  & --  &  0.72, 0.72   \\ \hline
\end{tabular}
}
\label{tab:classification-table1}
\end{table}

\begin{table}[t]
\caption{F1-macro, F1-micro for movie review dataset classification using K-NN with K=3 ($r_1=1, r_2=1$ is the \jose \, score).}
\resizebox{\columnwidth}{!}{
\begin{tabular}{|l|l|l|l|l|}
\hline
$r_1$\textbackslash{}$r_2$ & $r_2$=1 & $r_2$=2 & $r_2$=3 & $r_2$=4 \\ \hline
$r_1$=1              &   0.74, 0.74 & 0.75, 0.75   & 0.76, 0.76  & 0.75, 0.76  \\ \hline
$r_1$=2              &  --  & 0.75, 0.75 & 0.75, 0.75  & 0.74, 0.74  \\ \hline
$r_1$=3              &  --  &  --  & 0.74, 0.74 & \textbf{0.76, 0.76} \\ \hline
$r_1$=4              & --  & --  & --  & 0.74, 0.74\\ \hline
\end{tabular}
}
\label{tab:classification-table2}
\end{table}
\subsection{Semantic Textual Similarity Task}
Semantic Textual Similarity Benchmark comprises a selection of the English datasets used in the STS tasks organized in the context of SemEval \cite{Cer_2017} between 2012 and 2017\footnote{\url{http://ixa2.si.ehu.eus/stswiki/index.php/STSbenchmark}}. 
We perform semantic textual similarity tasks on the sts-benchmark dataset to show that even sentences can benefit from being represented as matrices. The benchmark comprises of 8\,628 sentence pairs split into 3 partitions: train, development and test. The results are reported on both the test and dev sets. Each sentence in the dataset is treated as a separate document by our model and we use all the sentences in the train, development and test set to train. The rationale for this is that the model is completely unsupervised, i.e., it takes only the raw text and uses no supervised or annotated information, and thus there is no need to hold out the test data as it is unlabelled. We train for 1\,000 iters with window size 15 and negative samples 5 while the rest of the hyperparameters were kept at their default values. 
To score a sentence pair representation, similarity was computed between them using our custom metric described in \ref{eq:1} for our model. We report the dev and test Pearson correlation score for $r_1, r_2=1,2,3,4, \, r_1 \leq r_2$. As Table \ref{tab:sts-table} reports, higher values of $r_2$ give better scores compared to our baseline model \jose \, ($r_1=r_2=1$).

\begin{table}[t]
\caption{Pearson Correlation for STS Benchmark on dev and test data ($r_1=1, r_2=1$ is the \jose \, score). Even sentences can benefit from our matrix representation as demonstrated by better scores with higher values of $r_2$.} 
\resizebox{\columnwidth}{!}{
\begin{tabular}{|l|l|l|l|l|}
\hline
$r_1$\textbackslash{}$r_2$ & $r_2$=1 & $r_2$=2 & $r_2$=3 & $r_2$=4 \\ \hline
$r_1$=1              &  0.51, 0.40  &  0.51, 0.39  &  0.52, 0.40  & 0.53, 0.40  \\ \hline
$r_1$=2              &  --  &  0.53, 0.40  & 0.53, 0.40  & 0.53, 0.40  \\ \hline
$r_1$=3              &  --  &  --  &  0.53, 0.40  & \textbf{0.54, 0.40}  \\ \hline
$r_1$=4              & --  & --  & --  &  0.53, 0.40  \\ \hline
\end{tabular}
}
\label{tab:sts-table}
\end{table}


\section{Conclusion}
In this paper, we extend the joint modelling idea used for training text embeddings from vectors with unit norm to matrices with unit Frobenius norm. Each word/sentence/document matrix is made to encode information in a way that each column of the matrix represents some latent topic, context, or discourse. Since the standard vector dot product can no longer be applied, we introduce a novel similarity metric that allows the measurement of similarity between matrices of arbitrary number of columns. For optimization simplicity, we reshape our matrices to vectors of unit norm that allows using the Riemannian gradient descent optimization algorithm on the spherical manifold. Our theory is validated quantitatively by the results which shows that our text embeddings outperform or produce similar results when compared with \jose \, in document classification, clustering, and semantic textual similarity tasks. 

This paper is meant to serve as a ground work for more involved research topics which integrate concepts of differential geometry and NLP. Future directions could include qualitative analysis on the columns of the matrices to see what tangible information they encode which will allow for better modelling. Another direction would be to exploit other matrix structures on the embeddings, e.g., treating each word embedding as a symmetric positive definite matrix and to study whether they can be beneficial. 

\bibliography{word_matrix}
\bibliographystyle{acL_natbib}

\appendix



\end{document}